\documentclass{article}
\usepackage[nonatbib,preprint]{nips_2018}

\usepackage{cite}

\usepackage[utf8]{inputenc} 
\usepackage[T1]{fontenc}    
\usepackage{hyperref}       
\usepackage{url}            
\usepackage{booktabs}       
\usepackage{amsfonts}       
\usepackage{nicefrac}       
\usepackage{microtype}      

\usepackage{graphicx}
\graphicspath{ {images/} }

\usepackage{algorithm}
\usepackage{algorithmic}

\def\vect#1{\mbox{\boldmath $#1$}}

\usepackage{wrapfig}

\title{Concept Bottleneck Model with Additional Unsupervised Concepts}

\author{
  Yoshihide Sawada\\
  Tokyo Research Center, Aisin\\
  \texttt{yoshihide.sawada@aisin.co.jp} \\
  \And
  Keigo Namamura\\
  Tokyo Research Center, Aisin\\
  \texttt{keigo.nakamura@aisin.co.jp} \\
}

\begin{document}

\maketitle

\begin{abstract}
With the increasing demands for accountability, interpretability is becoming an essential capability for real-world AI applications. However, most methods utilize post-hoc approaches rather than training the interpretable model. In this article, we propose a novel interpretable model based on the concept bottleneck model (CBM). CBM uses concept labels to train an intermediate layer as the additional visible layer. However, because the number of concept labels restricts the dimension of this layer, it is difficult to obtain high accuracy with a small number of labels. To address this issue, we integrate supervised concepts with unsupervised ones trained with self-explaining neural networks (SENNs). By seamlessly training these two types of concepts while reducing the amount of computation, we can obtain both supervised and unsupervised concepts simultaneously, even for large-sized images. We refer to the proposed model as the {\it concept bottleneck model with additional unsupervised concepts} ({\it CBM-AUC}). We experimentally confirmed that the proposed model outperformed CBM and SENN. We also visualized the saliency map of each concept and confirmed that it was consistent with the semantic meanings.
\end{abstract}

\section{Introduction}
\label{sec:introduction}
The lack of explainability of black-box neural networks is a serious barrier to the practical application of AI technologies. Especially when accountability is required, such as in medical and self-driving systems, black-box models may be expected to cause severe problems. To deploy more AI into the real world, neural networks are required to have greater interpretability.

Various methods have been developed to improve interpretability, and these can be categorized into two approaches, including (1) post-hoc and (2) ante-hoc. The first approach visualizes the features that are important for the trained network to determine its output. Although these methods are model-agnostic, they can be unreliable because they do not always contain the desired information~\cite{Sanitiy2018Adebayo}. In contrast, the second approach embeds human-understandable domain knowledge into the models. By incorporating this knowledge during training, models can prevent distortions in interpretation. 

One interpretable model is called the concept bottleneck model (CBM)~\cite{koh2020concept}. CBM has an intermediate layer representing human-understandable concepts as the additional visible layer (we call it {\it concept layer}). CBM outputs through this layer and trains it by supervised learning. After training, each unit in the concept layer outputs a single concept necessary for solving the target task. Thanks to this concept layer, CBM has high interpretability. However, CBM involves a constraint that the number of concepts limits the dimension of the concept layer. This constraint makes it difficult for CBM to achieve high accuracy with relatively small number of concept labels.

To solve this problem, we propose a novel interpretable model inspired by human judgement and decision-making. It is well known that humans do not simply make decisions based on the concepts (reasons) that we verbalize in our explanations. In other words, we use {\it implicit knowledge (know-how) in combination with explicit knowledge}~\cite{Polanyi}. However, the concept layer in CBM uses only explicit knowledge (supervised concepts). From this perspective, we extend this layer to enable the model to leverage explicit and implicit knowledge to perform task prediction, similar to humans.

We assume that implicit knowledge corresponds to the unsupervised concepts in the sense that they are not labeled (not verbalized). Namely, in order to perform the above extension, supervised and unsupervised concepts must be learned simultaneously. 

In this study, we adopt the self-explaining neural network (SENN) to obtain the unsupervised concepts. SENN is composed of an encoder-decoder architecture and a parametrizer, which estimates the weights of each concept~\cite{alvarez2018towards}. This method has been shown to be effective on the MNIST~\cite{lecun1998gradient} and CIFAR-10~\cite{krizhevsky2009learning} datasets. However, it is known that SENN is difficult to apply to the datasets of large images~\cite{koh2020concept}. Therefore, simply integrating SENN and CBM yields a model that is difficult to apply to complex problems such as self-driving systems. To address this issue, we adopt a weight-sharing technique and a discriminator, as an alternative to the decoder. By using these techniques, not only SENN but also our integrated model can be trained on datasets with large images. We call this integrated model the {\it concept bottleneck model with additional unsupervised concepts} ({\it CBM-AUC}). Experimental results using large image datasets (CUB-200-2011~\cite{wah2011caltech} and BDD-OIA~\cite{xu2020explainable}) show that our model was effective. We also visualized the saliency maps of each concept and found that the concepts can be consistent with their semantic meanings. This result indicates that the statement by Margeloiu et al.~\cite{2021Ashman} is not necessarily valid. 

The main contributions of this study are summarized as follows:
\begin{itemize}
\item We propose the concept bottleneck model with additional unsupervised concepts (CBM-AUC) based on the CBM and SENN. 
\item We apply a discriminator and weight-sharing to perform training efficiently.
\item Experimental results demonstrate the effectiveness of the proposed model.
\end{itemize}

\begin{figure}[t]
\centering
\begin{minipage}{1.0\hsize}
\begin{center}
\includegraphics[width=1.0\linewidth]{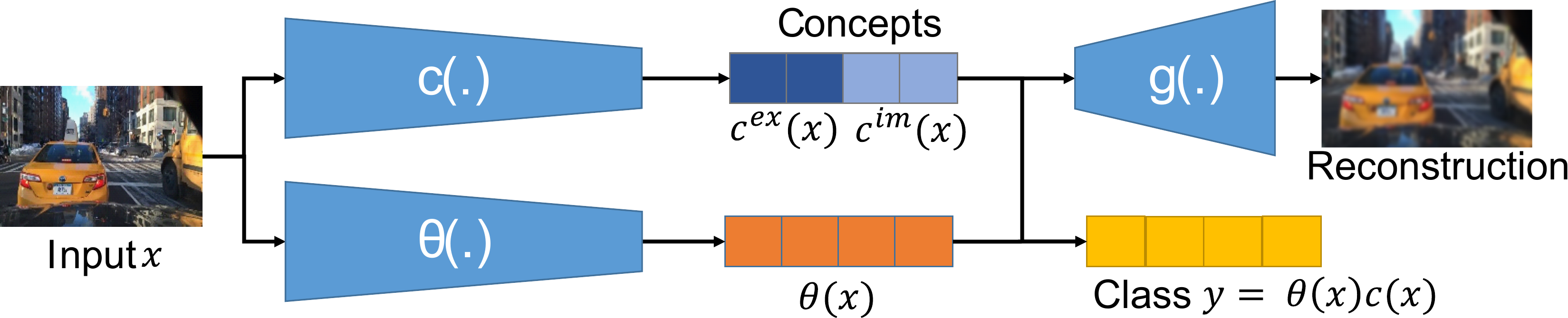}\\
(A) 
\end{center}
\end{minipage}
\begin{minipage}{1.0\hsize}
\begin{center}
\includegraphics[width=1.0\linewidth]{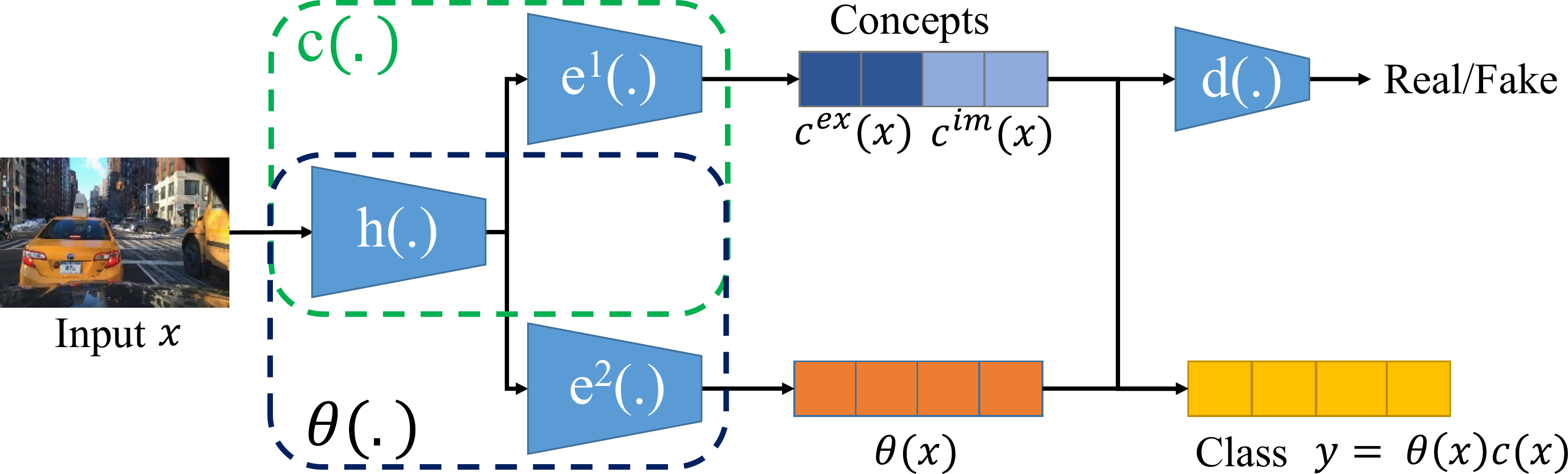}\\
(B)
\end{center}
\end{minipage}
\caption{Network overview of (A) a simple integration of CBM and SENN, and (B) our proposed CBM-AUC. $\vect{c}^{ex}$ and $\vect{c}^{im}$ represent explicit (supervised) and implicit (unsupervised) concepts. $\vect{g}(.)$ in (A) represents the decorder, and $d(.)$ in (B) represents the discriminator. $\vect{c}(.)$ and $\vect{\theta}(.)$ represent the encoder and parametrizer, which estimates the weights of each concept. In CBM-AUC, we set $\vect{c}(.) = \vect{e}^1(\vect{h}(.))$ and $\vect{\theta}(.) = \vect{e}^2(\vect{h}(.))$.}
\label{fig:network}
\end{figure}

\section{Related Work}
\label{sec:related_work}
Research on interpretable concepts can be mainly divided into post-hoc and ante-hoc methods.

For the post-hoc approach, Goyal et al.~\cite{goyal2019explaining} used a variational autoencoder to measure the causal concept effect of a trained model. Zhou et al.~\cite{zhou2018interpretable} decomposed saliency maps generated by CAM~\cite{zhou2016learning} or grad-CAM~\cite{selvaraju2017grad} to each concept, and Kim et al.~\cite{kim2018interpretability} and Ghorbani et al.~\cite{ghorbani2019towards} computed concept sensitivity using directional derivatives. Also, Bau et al.~\cite{bau2017network} used pixel-wise segmentation to estimate the concepts. These methods focused on the concept visualization of state-of-the-art models. However, there are no guarantees that these models use the desired concepts.

For ante-hoc approaches, Koh et al.~\cite{koh2020concept} proposed a CBM that predicted target tasks through the concept layer. Bahadori and Heckerman~\cite{bahadori2020debiasing} extended this model to denoise the concepts, and Bel{\'{e}}m et al.~\cite{2021Belem} applied a CBM to perform weakly supervised learning. Alvarez and Jakkola~\cite{alvarez2018towards} proposed a SENN to generalize the interpretable linear model and combines the encoder-decoder architectures as in~\cite{li2017deep}, and Marcinkevics and Vogt~\cite{2021granger} extended SENN for sequential data. 

Aside from these models, Losch et al.~\cite{losch2019interpretability} added a concept bottleneck layer to the trained segmentation network, and Kim et al.~\cite{kim2020visual} proposed a network architecture for visual concepts to capture the global context. Chen et al.~\cite{chen2020concept} proposed concept whitening as a normalization technique, and Li et al.~\cite{li2017deep} and Chen et al.~\cite{2018ChenThis} introduced an intermediate layer, which was not easy to train~\cite{2018Rudinstop} but which was designed to describe prototypes. Similarly, Cao et al.~\cite{2020CaoConcpt} proposed a meta-learning method that learned a mapping of concepts into semi-structured metric spaces and combined the outputs of each concept learner. These concept whitening and prototyping approaches require one additional patch image per concept.

Multi-task or auxiliary learning methods also use concepts to improve the target performance~\cite{huang2016part,xu2020explainable}. However, in these methods, the concepts may not affect the targets tasks~\cite{koh2020concept}.

\section{Method}
\label{sec:ours}
In this section, we first summarize existing two models, CBM and SENN. Then, we explain CBM-AUC in detail after describing two techniques to enable SENN to handle large-sized images.

\subsection{Concept Bottleneck Model (CBM)}
\label{sub:cbm}
CBM predicts target labels $y \in \mathbb{R}$ based on concepts $\vect{c} \in \mathbb{R}^{D_{\rm{ex}}}$, where $D_{\rm{ex}}$ is the number of supervised concepts~\cite{koh2020concept}. They proposed three training approaches: independent, sequential, and joint training. The independent approach first trains $\vect{x} \rightarrow \vect{c}$, then trains $\vect{c} \rightarrow y$ by using the ground truth concepts. The sequential approach also trains $\vect{x} \rightarrow \vect{c}$ at first, then trains $\vect{c} \rightarrow y$ by using the predicted concepts, not the ground truth concepts. In contrast, the joint approach trains $\vect{x} \rightarrow \vect{c} \rightarrow y$ simultaneously. We focus on joint training due to higher accuracy than other approaches~\cite{koh2020concept}.

Given $\{ \vect{x}_i, \vect{c}_i, y_i \}_{i=1}^N$, CBM is minimized as follows.
\begin{equation}
\mathcal{L}_{\rm{cbm}} = \sum_i^N \Bigl( \mathcal{L}_y(f(\vect{x}_i),y_i) +  \lambda \sum_j^{D_{\rm{ex}}} \mathcal{L}_j(c_j^{\rm{ex}}(\vect{x}_i), c_{i,j}) \Bigr),
\end{equation}
where $\mathcal{L}_y$ represents the classification loss, $\mathcal{L}_j$ represents the $j$-th concept loss, $f(\vect{x})$ and $c_{j}^{\rm{ex}}(\vect{x})$ are the predictions of targets and $j$-th concept, $c_{i,j}$ is the $j$-th ground truth concept of $\vect{x}_i$, $\vect{c}_i = [c_{i,1}, c_{i,2}, \cdots, c_{i, D_{\rm{ex}}}]^\top$, and $\lambda$ is a hyperparameter. These losses measure the discrepancy by the standard loss functions, e.g., cross-entropy and mean squared error.

\subsection{Self-Explaining Neural Network (SENN)}
\label{sub:senn}
SENN is an unsupervised concept learning method that uses an interpretable linear model~\cite{alvarez2018towards}. 
The prediction of SENN is described as
\begin{eqnarray}
f(\vect{x}) = \vect{\theta}(\vect{x})^\top \vect{c}^{\rm{im}}(\vect{x}), 
\end{eqnarray}
where $\vect{\theta}(.)$ is a parametrizer which estimates weights of each concept, and $\vect{c}^{\rm{im}}(.)$ is the encoder to output unsupervised concepts. $\vect{c}^{\rm{im}}(\vect{x}), \vect{\theta}(\vect{x}) \in \mathbb{R}^{D_{\rm{im}}}$, and $D_{\rm{im}}$ represents the number of dimensions of unsupervised concepts. SENN represents $\vect{c}^{\rm{im}}(.)$ and $\vect{\theta}(.)$ by neural networks, respectively. The main difference between SENN and CBM is that SENN trains networks without using concept labels (i.e., $\{\vect{x}_i, y_i\}_{i=1}^N$). Given the training dataset $\{\vect{x}_i, y_i\}_{i=1}^N$, SENN minimizes the following loss function.
\begin{equation}
\mathcal{L}_{\rm{senn}} = \sum_i^N \mathcal{L}_y(f(\vect{x}_i),y_i) +  \alpha \mathcal{L}_{x}(\vect{x}_i,\vect{g}(\vect{c}^{\rm{im}}(\vect{x}_{i}))) + \beta \mathcal{L}_{\theta}(f(\vect{x}_i)), 
\end{equation}
where $\mathcal{L}_y$ represents the classification loss, and $\mathcal{L}_{x}$ represents the reconstruction error. $\vect{g}(\vect{c}^{\rm{im}}(\vect{x}_{i}))$ is the reconstructed image using the decoder $\vect{g}(.)$, and $\alpha$ and $\beta$ are the hyperparameters. $\mathcal{L}_{\theta}$ is the regularization term representing the stability of $\vect{\theta}(\vect{x})$ by the gradient. 
\begin{equation}
\mathcal{L}_{\theta}(f(\vect{x})) = || \vect{\nabla}_x f(\vect{x}) - ( \vect{\theta}(\vect{x})^\top \vect{J}^{c^{\rm{im}}}_x )^\top || \approx 0,
\label{eqn:grad_senn}
\end{equation}
where $\vect{\nabla}_x f(\vect{x})$ is the derivative of $f(\vect{x})$ and $\vect{J}^{c^{\rm{im}}}_x$ is the Jacobian of the concept $\vect{c}^{\rm{im}}(\vect{x})$ with respect to $\vect{x}$. This regularization term makes $\vect{\theta}(\vect{x})$ robust to small changes in concepts \cite{alvarez2018towards}. 

\subsection{Adapting SENN to Large Images}
\label{sub:scaling}
In most autoencoder networks, including SENN, the same network architecture is used for the encoder and decoder. Therefore, the reconstruction error requires a large number of parameters~\cite{hafner2019dream}. In addition, SENN must compute the Jacobian $\vect{J}^{c^{\rm{im}}}_x \in \mathbb{R}^{D_{\rm{im}} \times D}$ with respect to each input $\vect{x} \in \mathbb{R}^D$. Therefore, as the input size increases, the size of the Jacobian to be computed also increases. 

We adopt two techniques to address this problem. First, we use the discriminator proposed by Hejelm et al.~\cite{hjelm2018learning} instead of the decoder~(see Sec.~\ref{sub:architecture} for details of the network architecture). To train this discriminator $d(.)$, we adopt a simple least squared error as follows.
\begin{equation}
\mathcal{L}_{\rm{dis}}(d(\vect{z}),d(\vect{z}^{\prime})) = || d(\vect{z}) - a ||^2 + || d(\vect{z}^{\prime}) - b ||^2,
\label{eqn:dis}
\end{equation}
where $\vect{z} = [\vect{c}^{\rm{im}}(\vect{x})^\top,\vect{h}(\vect{x})^\top]^\top$, $\vect{z}^\prime = [\vect{c}^{\rm{im}}(\vect{x})^\top,\vect{h}(\vect{x}^\prime)^\top]^\top$, $\vect{x}^\prime$ is the input differed from $\vect{x}$, and $\vect{h}(.)$ is the intermediate output before the concept layer. $\vect{z}$ and $\vect{z}^\prime$ represent real and fake~\cite{hjelm2018learning}, and $a$ and $b$ represent the labels of them. In this study, we set $a = 1$ and $b = 0$.

Second, we share the intermediate network $\vect{h}(\vect{x}) \in \mathbb{R}^{D_{\rm{h}}}$ with $\vect{c}^{\rm{im}}(\vect{x})$ and $\vect{\theta}(\vect{x})$. By sharing $\vect{h}(\vect{x})$, $\vect{\nabla}_x f$ and $\vect{J}^{c^{\rm{im}}}_x$ can be transformed by the chain rule as follows.
\begin{eqnarray}
\vect{\nabla}_x f(\vect{x}) &=& ( \vect{\nabla}_h f(\vect{x})^\top \vect{J}^h_x )^\top,\\
\vect{J}^{c^{\rm{im}}}_x &=& \vect{J}^{c^{\rm{im}}}_h \vect{J}^h_x.
\end{eqnarray}
By substituting the these equations to Eq.(\ref{eqn:grad_senn}), we obtain
\begin{eqnarray}
\tilde{\mathcal{L}}_{\theta}(f(\vect{x})) = || \vect{\nabla}_h f(\vect{x}) - ( \vect{\theta}(\vect{x})^\top \vect{J}^{c^{\rm{im}}}_h )^\top || \approx 0.
\end{eqnarray}
By using this equation, we need only compute $\vect{J}^{c^{\rm{im}}}_h$, the size of $D_{\rm{im}} \times D_{\rm{h}}$. Therefore, the computational burden is reduced under $D > D_{\rm{h}}$. Furthermore, we can use sophisticated pretrained models (e.g., Inception-v3~\cite{szegedy2016rethinking}) as the intermediate layer $\vect{h}(.)$ regardless of the decoder.

From these techniques, the modified loss function of SENN is as follows. 
\begin{equation}
\tilde{\mathcal{L}}_{\rm{senn}} = \sum_i^N \mathcal{L}_y(f(\vect{x}_i),y_i) +  \alpha \mathcal{L}_{\rm{dis}}(d(\vect{z}_i),d(\vect{z}_i^{\prime})) + \beta \tilde{\mathcal{L}}_{\theta}(f(\vect{x}_i)). 
\end{equation}
We use this loss function because the original SENN could not learn the dataset used in our experiments. We call this model the modified SENN (M-SENN).

\subsection{Concept Bottleneck Model with Additional Unsupervised Concepts (CBM-AUC)}
\label{sec:cbm-auc}
As described in Sec.~\ref{sec:introduction}, because the prediction of the CBM has a close link to explicit knowledge, the model becomes a highly interpretable DNN model. However, CBM has a constraint that the number of concepts limits the concept layer's dimension. This constraint leads to low classification performance. We address this problem by adding unsupervised concepts (implicit knowledge), i.e., combining the CBM and M-SENN as follows.
\begin{eqnarray}
\mathcal{L}_{\rm{cbmauc}} = \sum_i^N \Bigl( \mathcal{L}_y(f(\vect{x}_i),y_i) + \alpha \mathcal{L}_{\rm dis}(d(\vect{z}_i),d(\vect{z}_i^{\prime})) \nonumber\\
+ \beta \tilde{\mathcal{L}}_{\theta}(f(\vect{x}_i)) + \lambda \sum_j^{D_{\rm{ex}}} \mathcal{L}_j(c_j^{\rm{ex}}(\vect{x}_i), c_{i,j}) \Bigr).
\label{eqn:ss-sennd}
\end{eqnarray}
The first term corresponds to classification loss, and the second and third terms correspond to the M-SENN's losses. The last term corresponds to the CBM's regularization for the supervised concepts. Note that we compute $\tilde{\mathcal{L}}_{\theta}$, $\mathcal{L}_{\rm dis}$, and $\mathcal{L}_y$ over all concepts $\vect{c}(\vect{x}) = [(\vect{c}^{{\rm{ex}}}(\vect{x}))^\top, (\vect{c}^{{\rm{im}}}(\vect{x}))^\top]^\top \in \mathbb{R}^{D_{\rm{ex}}+D_{\rm{im}}}$, therefore, $f(\vect{x})$ becomes as follows.
\begin{eqnarray}
f(\vect{x}) = \vect{\theta}(\vect{x})^\top \vect{c}(\vect{x}) = \vect{e}^2(\vect{h}(\vect{x}))^\top \vect{e}^1(\vect{h}(\vect{x})), \end{eqnarray}
where $\vect{e}^1$ and $\vect{e}^2$ are the remainder of networks. An overview of the proposed network  is shown in Fig.~\ref{fig:network}. We also show a network integrating the original SENN and CBM for ease of comparison.

\section{Experimental Setup}
\label{sec:setting}
In this study, we evaluated our method on the CUB-200-2011~\cite{wah2011caltech} and BDD-OIA~\cite{xu2020explainable} datasets. For these experiments, we used one GPU, an NVIDIA Tesla v100, with 32GB.

\subsection{Datasets}
\label{sec:dataset}
CUB-200-2011 (Caltech-UCSD Birds-200-2011) is dataset of 11788 images of birds (training: 4796; validation: 1198; test: 6794) which have 200 species and 312 concepts. In Koh et al.~\cite{koh2020concept}, the authors used the 112 denoising binary bird concepts (e.g., ``Wing color'', ``Beak shape''), and we followed their configuration. 

BDD-OIA consists of 22924 videos (training: 16802; validation: 2270; test: 4572), which have four actions (``Forward'', ``Stop'', ``Left'', ``Right'') and 21 concepts (e.g., ``Traffic light is green'', ``Load is clear''). In this study, we used only the last frame of each video as in the original paper~\cite{xu2020explainable}. Note that each image has multiple actions, differing from CUB-200-2011. 

\subsection{Network Architecture}
\label{sub:architecture}
For the CUB-200-2011, we used the Inception-v3 network~\cite{szegedy2016rethinking} pretrained by the ImageNet as the intermediate network $\vect{h}$, as in Koh et al.~\cite{koh2020concept}. Following $\vect{h}$, we set $\vect{c}$ and $\vect{\theta}$. As $\vect{c}$, we used a fully connected (FC) layer to output $D_{\rm{ex}} + D_{\rm{im}}$ concepts. As $\vect{\theta}$, we used three FC layers ($2048 \times 1024$, $1024 \times 512$, $512\times (D_{\rm{ex}}+D_{\rm{im}})$) because $\vect{\theta}$ required more complex network than $\vect{c}$~\cite{alvarez2018towards}. For the first two layers of $\vect{\theta}$, we used batch normalization before the activation function Mish~\cite{misra2019mish}. As the discriminator $d$, we set the three FC layers ($(D_{\rm{ex}}+D_{\rm{im}}) \times 512$, $512 \times 512$, and $512 \times 1$, the same as global deep infomax architecture  in~\cite{hjelm2018learning}), and the activation of the first two layers is Mish.

For the BDD-OIA, we used Faster RCNN network~\cite{frcnn} as $\vect{h}$, as in Xu et al.~\cite{xu2020explainable}. This network was pretrained by COCO~\cite{coco}, fine-tuned with BDD100k~\cite{yu2020bdd100k}, and was frozen during training. Following Faster RCNN, we computed the input features for $\vect{c}$ and $\vect{\theta}$ by referring to the global module~\cite{xu2020explainable}. The difference is that we flattened the output of the module and executed one FC layer with ReLU to match the input of $\vect{c}$ and $\vect{\theta}$ described above. Note that the network architectures of $\vect{c}$, $\vect{\theta}$, and $d$ are the same as those used on the CUB-200-2011 experiment.

It should be noted that the selection of $\vect{h}(.)$ is arbitrary as long as $\vect{h}(.)$ is a good representation of the input due to the desiderata of SENN~\cite{alvarez2018towards}. Assuming that the well-known pre-trained models are satisfied with this condition, we selected the pre-trained Inception-v3 and faster RCNN for CUB200-2011 and BDD-OIA, respectively, as in \cite{koh2020concept} and \cite{xu2020explainable}.

\subsection{Hyperparameters}
According to Alvarez and Jakkola~\cite{alvarez2018towards}, sparsity is the essential property for unsupervised concepts. In this study, we used the $k$-WTA \cite{kWT2019AXiao} activation for the unsupervised concepts. $k$-WTA retained the $k$ largest values ($k=0,1,2,\cdots,D_{\rm{im}}$) and set all others to be zero. This is a natural extension of Winner-Take-All, which has been widely studied in the field of the spiking neural networks~\cite{diehl2015unsupervised}. By using this activation, it is unnecessary to add the sparsity regularization (e.g., L1 norm~\cite{alvarez2018towards}) 
such that the hyperparameter is real-valued. We set $k$ so that 50\% of unknown concepts were zero for CUB-200-2011 and $k=7$ for BDD-OIA, while maintaining a balance between accuracy and sparsity.

Followings are the other hyperparameters. For the CUB-200-2011, we used the cross-entropy and MSE losses for $\mathcal{L}_y$ and $\mathcal{L}_j$, respectively. We set $D_{\rm{ex}} + D_{\rm{im}} = 256$, and the batch size was $64$. As the implementation of Koh et al.~\cite{koh2020concept,koh2020git}, each model was trained three times over random seeds and computed the mean $\pm$ 2.0 standard deviation. For the BDD-OIA, we used the BCE loss for $\mathcal{L}_y$ and $\mathcal{L}_j$. In addition, we set $D_{\rm{ex}} + D_{\rm{im}} = 30$, and the batch size was $16$. In all experiments, the epoch was set to 50. Note that these hyperparameters, including comparison methods, are properly determined beforehand by the experiments.

We further examined the optimal hyperparameters for both datasets as follows. Optimizer: \{SGD, Adam\}, learning rate: $\{ 0.001, 0.01 \}$, $\lambda$: $\{ 0.1, 0.5, 1.0, 2.0, 3.0 \}$, $\alpha$: $\{ 0.1, 0.5, 1.0 \}$, and $\beta$: $\{ 0.001, 0.01\}$. For the other hyperparameters, we used the default values of PyTorch~\cite{paszke2019pytorch}. 

\section{Experimental Results}
\label{sec:result}

\subsection{Effectiveness of Network Sharing and Discriminator}

\begin{table}[t]
\centering
\begin{tabular}{ccc}
\hline
Dataset & $D_{h}/D$ & Parameter reduction \\
\hline\hline
CUB-200-2011 & 0.014 & 0.404 \\
BDD-OIA & $7.407\times10^{-4}$ &0.377  \\
\hline
\end{tabular}
\caption{Reduction effects of network sharing and discriminator. Parameter reduction was obtained by dividing the number of network parameters of simple integration of CBM and original SENN.}
\label{tbl:param_red}
\end{table}

We first show the effectiveness of the network sharing and discriminator. Table~\ref{tbl:param_red} shows the result of integration with the case where the simple combination was conducted. Here, $D_{h}/D$ and parameter reduction represent the reduction rate of the size of the Jacobian and the network parameters, respectively.

As shown in this table, we were able to significantly reduce the size of the Jacobian ($\mathcal{O}(10^{-2})$ on CUB-200-2011, $\mathcal{O}(10^{-4})$ on BDD-OIA). In addition, the number of parameters was reduced to about 40\%. As described in Sec.~\ref{sec:ours}, original SENN and simple integration could not run on CUB-200-2011 and BDD-OIA. In addition, the other variations could not run either. In contrast, M-SENN and our CBM-AUC were able to function in our environment. Therefore, we only show the results of the M-SENN and our CBM-AUC in the following experimental results.

\subsection{Experimental Results on CUB-200-2011}
\label{sub:result_cub}
In this section, we show the experimental results on CUB-200-2011. Following Koh et al.~\cite{koh2020concept}, we evaluated the performance by an average 0-1 task error and a root mean squared error (RMSE) of supervised concepts. We also used a linear probe~\cite{kim2018interpretability,koh2020concept} to compute the concept's RMSE of SENN because there was no one-to-one correspondence between the unsupervised concept and the unit.

\subsubsection{Performance Comparison}
\label{subsub:cub}

\begin{table}[t]
\centering
\begin{tabular}{ccc}
\hline
Model & Task & Concept\\
\hline \hline
CBM &  0.199 $\pm$ 0.006 & 0.031 $\pm$ 0.000\\
M-SENN & 0.191 $\pm$ 0.003 & 0.053 $\pm$ 0.004\\
CBM-AUC & 0.177 $\pm$ 0.009 & 0.028 $\pm$ 0.001\\
\hline
\end{tabular}
\caption{Comparison results for CUB-200-2011.}
\label{tbl:cub}
\end{table}

We compared CBM-AUC with CBM~\cite{koh2020concept} and M-SENN. For a fair comparison, we used 256 unsupervised concepts of M-SENN so that the total number matched CBM-AUC. In contrast, CBM used only 112 supervised concepts because it cannot be increased beyond $D_{\rm{ex}}$. Note that this is clearly a situation in line with our motivation (see Sec.~\ref{sec:introduction}). See Fig.~\ref{fig:cub_limit_concept} for the result when the total number of concepts is equal to the CBM. 

As shown in table~\ref{tbl:cub}, CBM-AUC outperformed CBM for task performance because CBM can only use supervised concepts. In addition, CBM-AUC also outperformed the concept accuracy of CBM and the task and concept accuracies of M-SENN. We consider that the higher accuracy of the concept than CBM was due to the disentanglement of explicit and implicit knowledge by the increased dimensionality of the concept layer. Therefore, using supervised and unsupervised concepts is necessary to obtain high performance.

\subsubsection{Performance when changing the number of concepts}
\label{sec:change_num_concept}

\begin{table}[t]
\centering
{\begin{tabular}{cccc}
\hline
Model & $D_{\rm{ex}}+D_{\rm{im}}$ & Task & Concept\\
\hline \hline
CBM-AUC & 128 & 0.183 $\pm$ 0.003 & 0.028 $\pm$ 0.001\\
CBM-AUC & 256 & 0.177 $\pm$ 0.009 & 0.028 $\pm$ 0.001\\
\hline
\end{tabular}}\\
\caption{CBM-AUC's performance comparison between $D_{\rm{ex}}+D_{\rm{im}}=256$ and $D_{\rm{ex}}+D_{\rm{im}}=128$.}
\label{tbl:cub_change_unknown}
\end{table}

\begin{figure}[t]
\centering
\begin{minipage}{0.465\hsize}
\begin{center}
\includegraphics[width=1.0\linewidth]{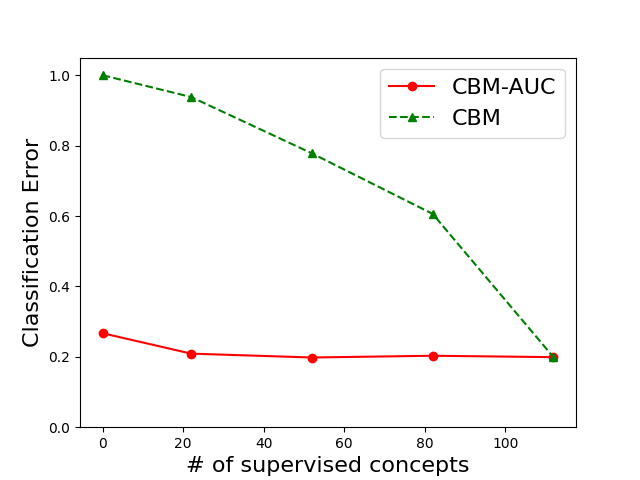}\\
(A)
\end{center}
\end{minipage}
\begin{minipage}{0.465\hsize}
\begin{center}
\includegraphics[width=1.0\linewidth]{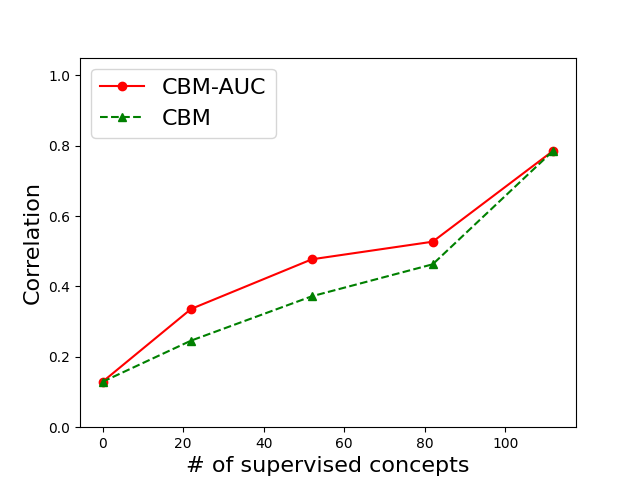}\\
(B)
\end{center}
\end{minipage}
\caption{Performance when limited supervised concepts. (A): change of the task performance, (B): change of the concept performance.}
\label{fig:cub_limit_concept}
\end{figure}

This section investigates the performance when the number of concepts $D_{\rm{ex}}$ and $D_{\rm{im}}$ changed. 

Table \ref{tbl:cub_change_unknown} shows the CBM-AUC's performances of $D_{\rm{ex}}+D_{\rm{im}}=256$ and $D_{\rm{ex}}+D_{\rm{im}}=128$. Note that the number of supervised concepts $D_{\rm{ex}}$ was fixed ($D_{\rm{ex}} = 112$). As shown in this table, the task performance decreased when $D_{\rm{im}}$ was small. In addition, CBM-AUC performed better than CBM (Table~\ref{tbl:cub}) when $D_{\rm{im}}=16$ ($D_{\rm{ex}}+D_{\rm{im}}=128$). This result indicates that the unsupervised concepts are important even when $D_{\rm{im}}$ is small.

We also investigated the performance when the number of supervised concepts was limited. We randomly removed the supervised concepts from the $D_{\rm{ex}} = 112$ and added the same number of unsupervised concepts, keeping the number of total concepts $D_{\rm{ex}}+D_{\rm{im}} = 112$ fixed. 

When the number of supervised concepts was reduced, RMSE could not accurately evaluate the concept accuracy. Therefore, we evaluated the concept correlation, defined as follows.
\begin{eqnarray}
\bar{r}^2 = \frac{1}{D_{\rm{ex}}} \sum_j \max_i r^2(c_j, c_i(\vect{x})), 
\end{eqnarray}
where $\bar{r}^2$ is the average coefficient, $r$ is the coefficient between two concepts, $c_{j}$ is the ground truth of supervised concept, and $c_i(\vect{x}) (i = 1, 2, \cdots, D_{\rm{ex}}+D_{\rm{im}})$ is the $i$-th predicted concept. Namely, $\bar{r}^2$ represents how the predicted concepts are close to the ground truth concepts. 

Figure~\ref{fig:cub_limit_concept} shows the task error and correlation of CBM and CBM-AUC. The red and green lines represent CBM-AUC and CBM, respectively. Note that $D_{\rm{ex}}=0$ and $D_{\rm{ex}}=112$ in CBM-AUC are equivalent to M-SENN and CBM. Comparing these results, the task performance of CBM decreased significantly when the concept correlation decreased, whereas the task performance of CBM-AUC changed only slightly. Meanwhile, as shown in Fig.~\ref{fig:cub_limit_concept} (B), CBM-AUC's correlation decreased similarly to CBM's one. If the additional unsupervised concepts acquire concepts similar to the supervised ones, the correlation should also be constant in CBM-AUC. This decreasing tendency implies that there was low correlation between supervised and unsupervised concepts. In future works, we intend to obtain the explicit concepts in an unsupervised manner.

\subsection{Experimental Results on BDD-OIA}
\label{sub:result_bdd}
In this section, we show the experimental results on BDD-OIA. Following Xu et al.~\cite{xu2020explainable}, we evaluated the performance by two F1 scores, $F1_{all}$ and $mF1$. $F1_{all}$ averages the F1 scores over all the predictions, and $mF1$ computes the mean of the F1 score for each action. We also computed the F1 scores of concept in the same manner.

\subsubsection{Performance Comparison}
\begin{table*}[t]
\centering
{\begin{tabular}{ccccccccc}
\hline
Model & F & S & R & L & $mF1$ & $F1_{all}$ & $mF1_{cpt}$ & $F1_{cpt,all}$ \\
\hline \hline
CBM & 0.795 & 0.732 & 0.431 & 0.483 & 0.610 & 0.661 & 0.292 & 0.412 \\
M-SENN & 0.705 & 0.727 & 0.339 & 0.385 & 0.539 & 0.612 & 0.098 & 0.216 \\
CBM-AUC & 0.803 & 0.751 & 0.525 & 0.551 & 0.658 & 0.704 & 0.342 & 0.522 \\
\hline
\end{tabular}}\\
\caption{Performance of the action and reason prediction for BDD-OIA. F/S/R/L represent the F1 scores of actions of ``Forward'', ``Stop'', ``Left'', and ``Right'', respectively. $mF1$ and $F1_{all}$ represent the F1 scores of actions, and $mF1_{cpt}$ and $F1_{cpt,all}$ represent the F1 scores of supervised concepts.}
\label{tbl:BDD}
\end{table*}

\begin{figure}[t]
\centering
\begin{minipage}{0.45\hsize}
\begin{center}
\includegraphics[width=1.0\linewidth]{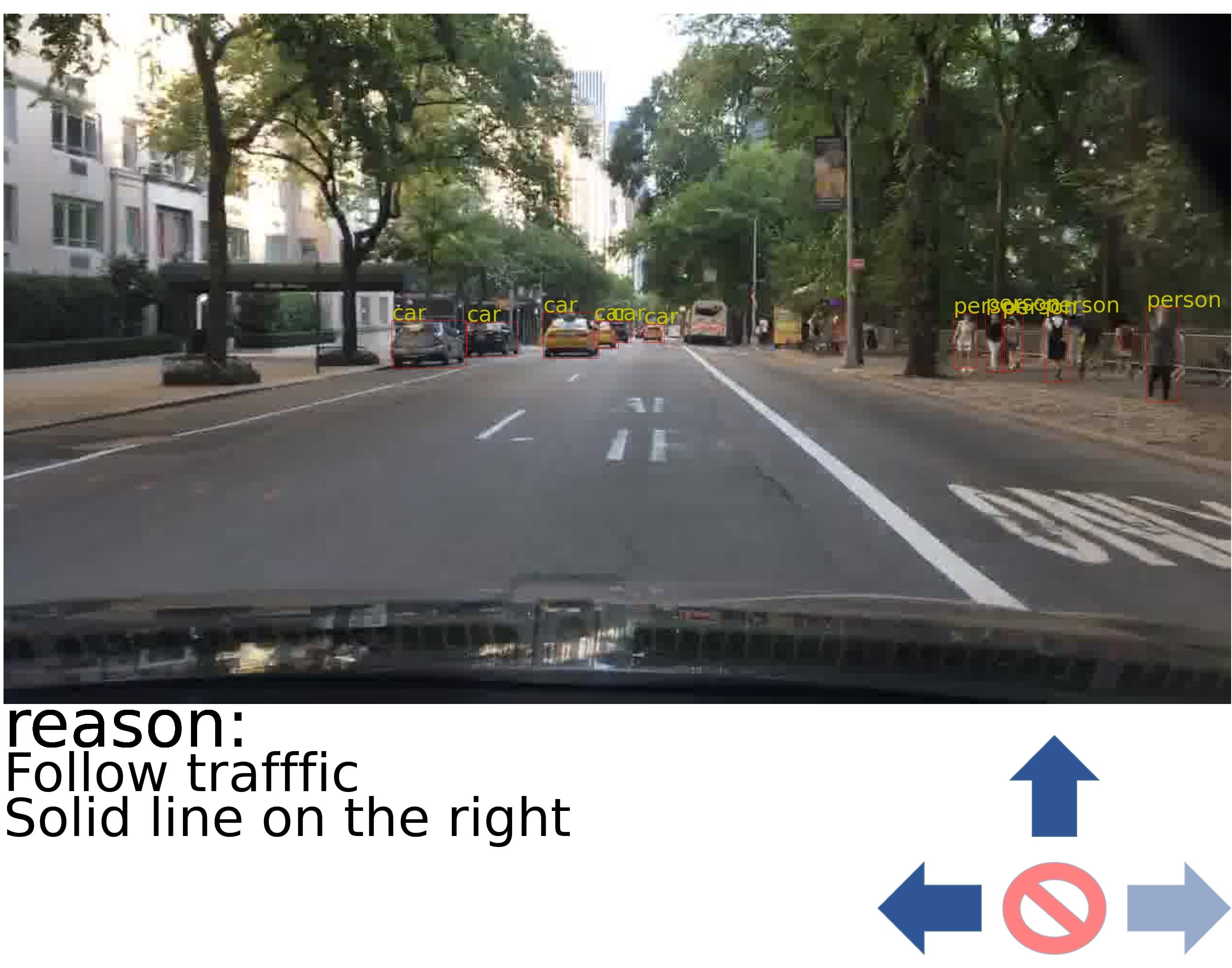}\\
(A)
\end{center}
\end{minipage}
\begin{minipage}{0.45\hsize}
\begin{center}
\includegraphics[width=1.0\linewidth]{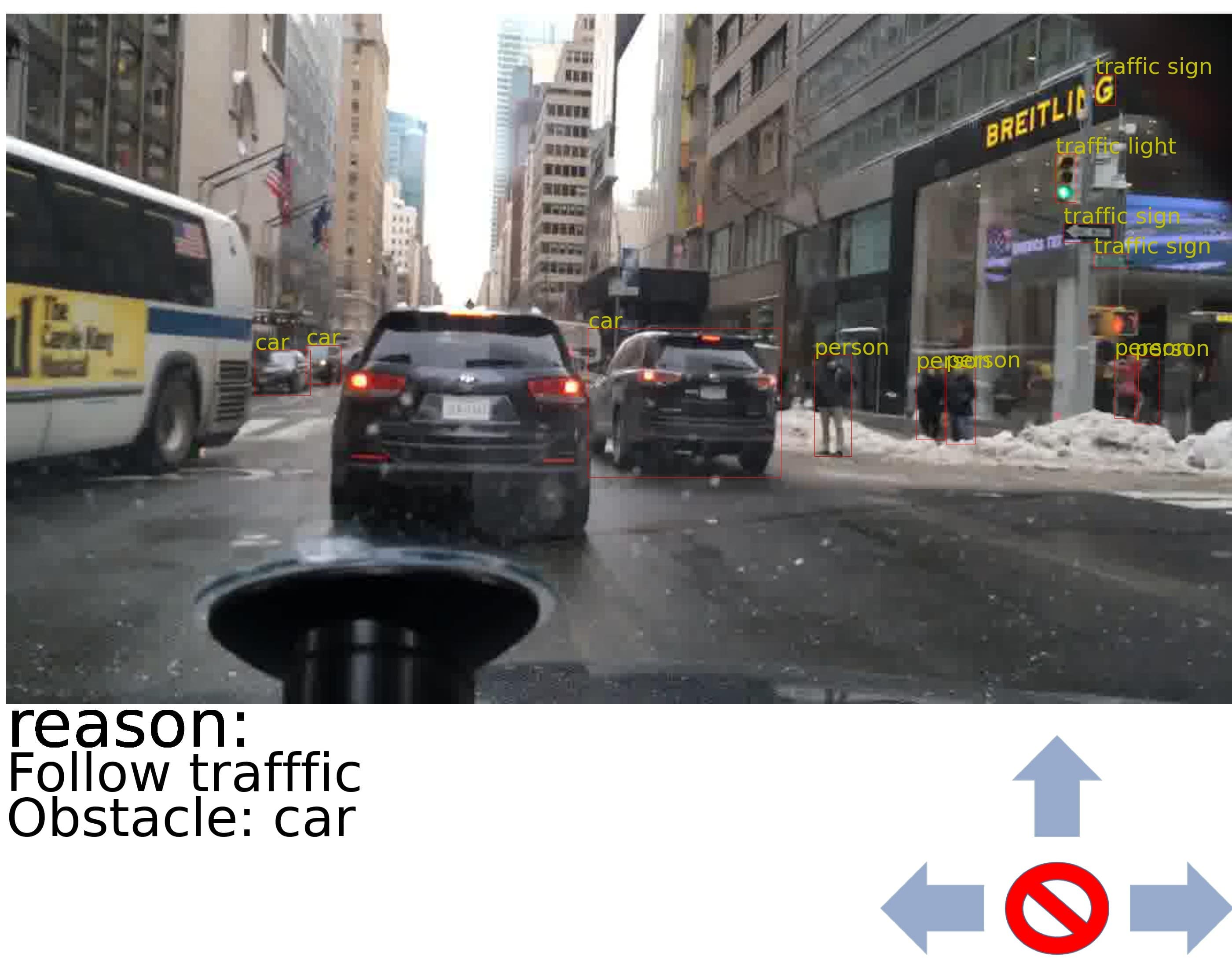}\\
(B)
\end{center}
\end{minipage}
\caption{The examples of the output of CBM-AUC. In these situations, our model can predict actions and concepts correctly.}
\label{fig:CBM-AUC_succeed}
\end{figure}

\begin{figure}[tb]
\centering
\begin{minipage}{1.0\hsize}
\begin{center}
\includegraphics[width=0.45\linewidth]{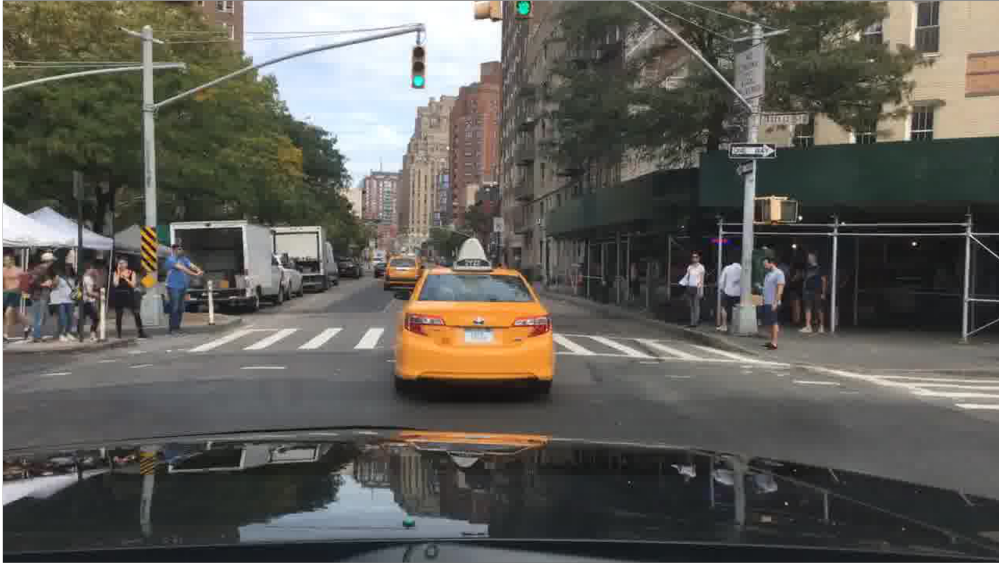}
\includegraphics[width=0.45\linewidth]{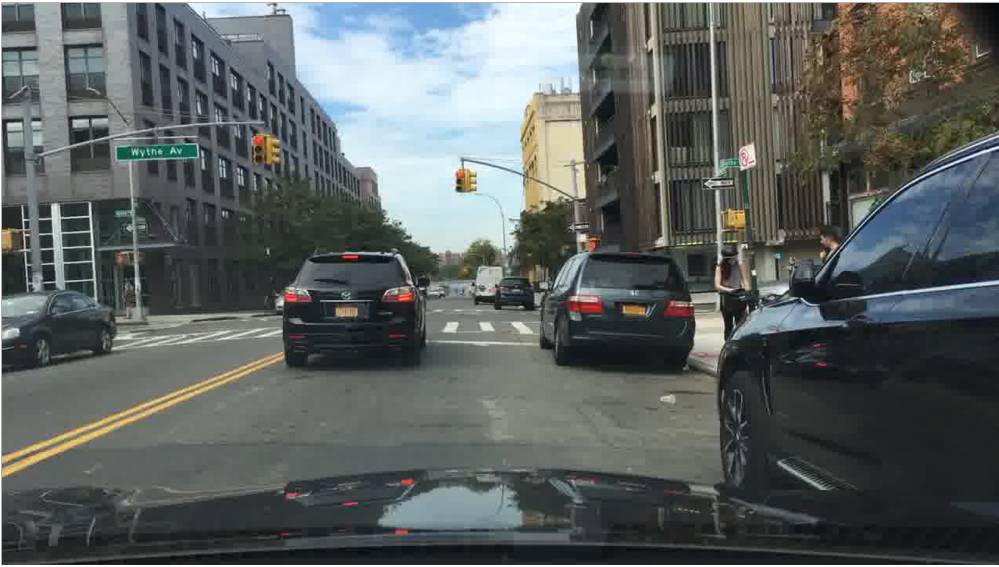}\\
(A)
\vspace{0.05cm}
\end{center}
\end{minipage}
\begin{minipage}{1.0\hsize}
\begin{center}
\includegraphics[width=0.45\linewidth]{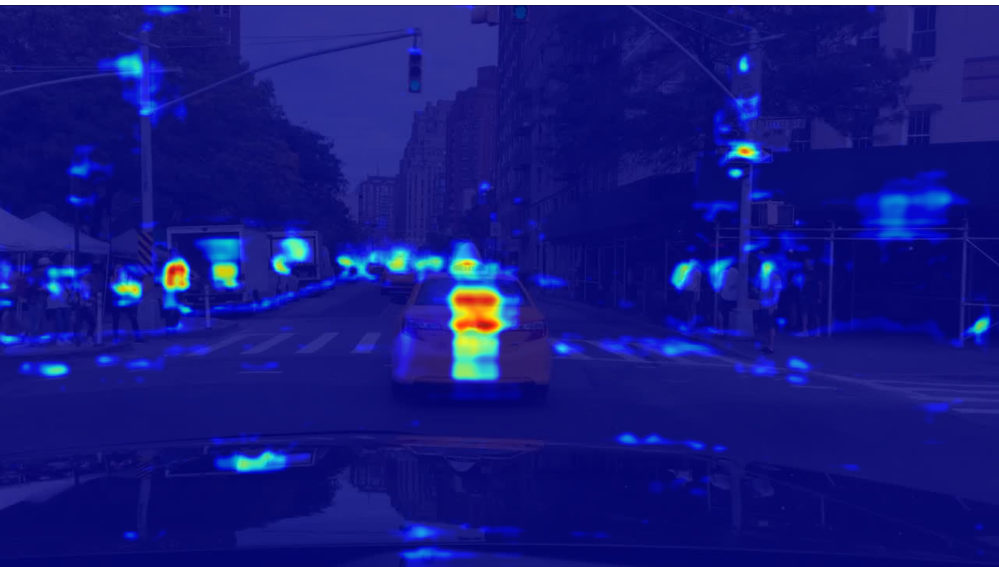}
\includegraphics[width=0.45\linewidth]{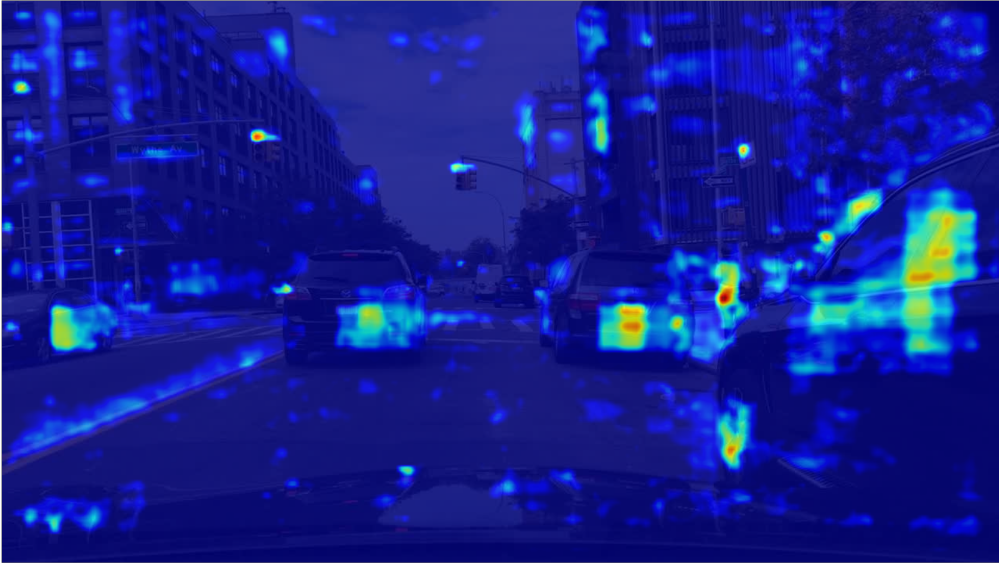}\\
(B)
\vspace{0.05cm}
\end{center}
\end{minipage}
\begin{minipage}{1.0\hsize}
\begin{center}
\includegraphics[width=0.45\linewidth]{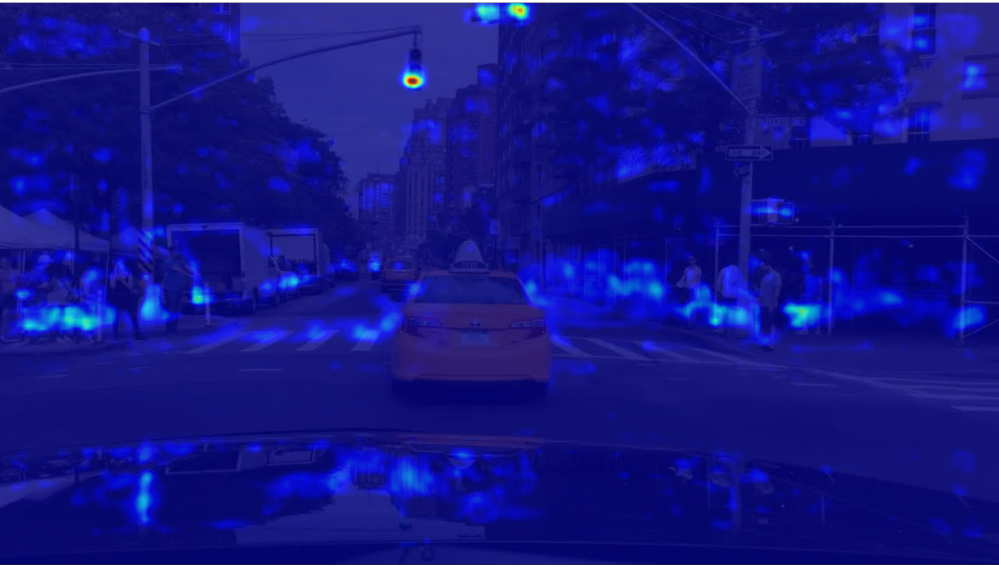}
\includegraphics[width=0.45\linewidth]{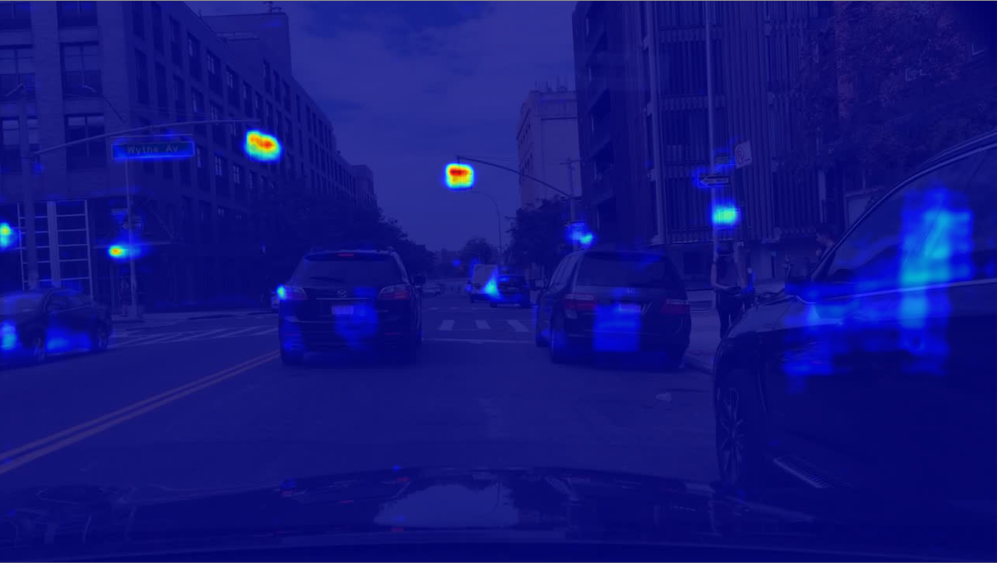}\\
(C)
\vspace{0.05cm}
\end{center}
\end{minipage}
\begin{minipage}{1.0\hsize}
\begin{center}
\includegraphics[width=0.45\linewidth]{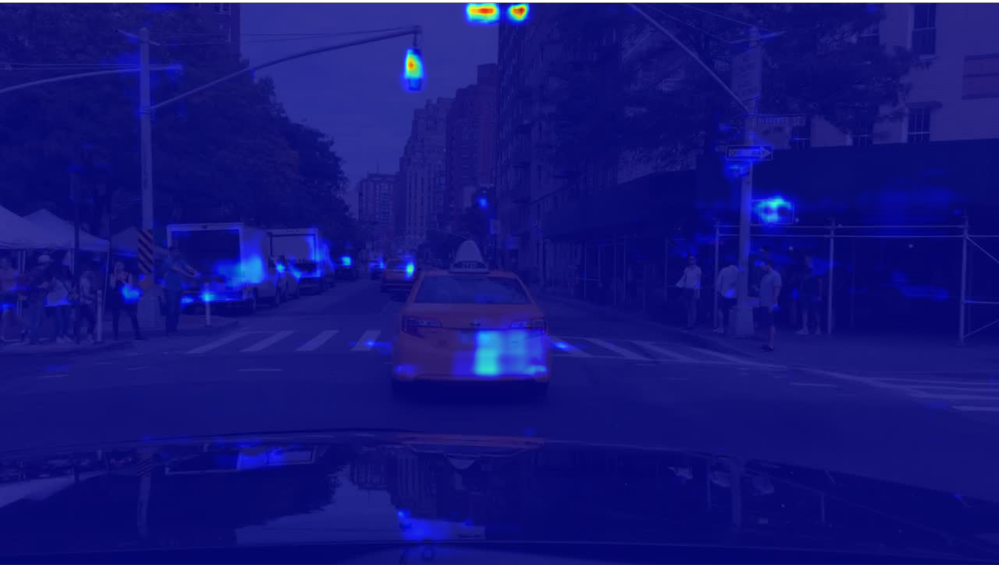}
\includegraphics[width=0.45\linewidth]{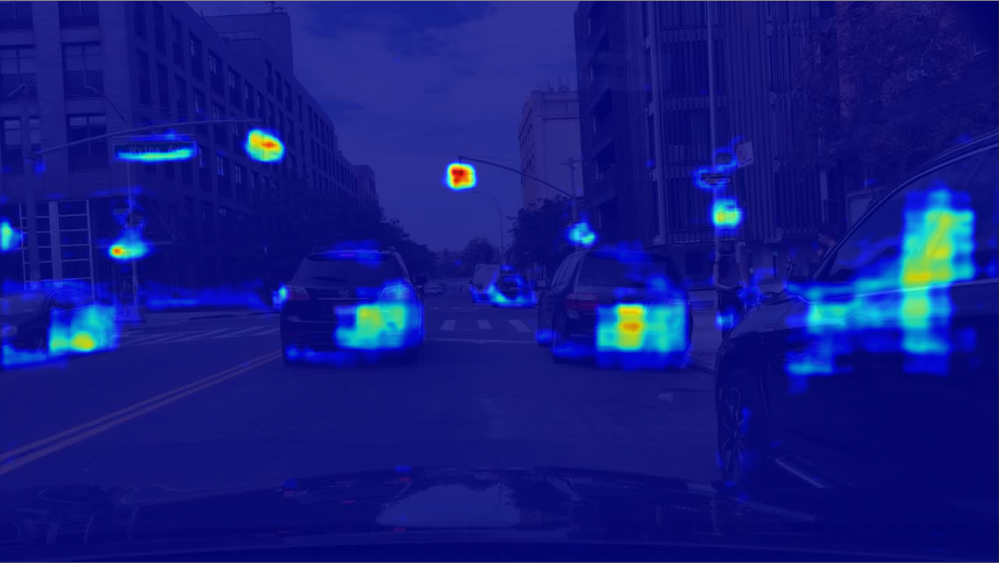}\\
(D)
\vspace{0.05cm}
\end{center}
\end{minipage}
\begin{minipage}{1.0\hsize}
\begin{center}
\includegraphics[width=0.45\linewidth]{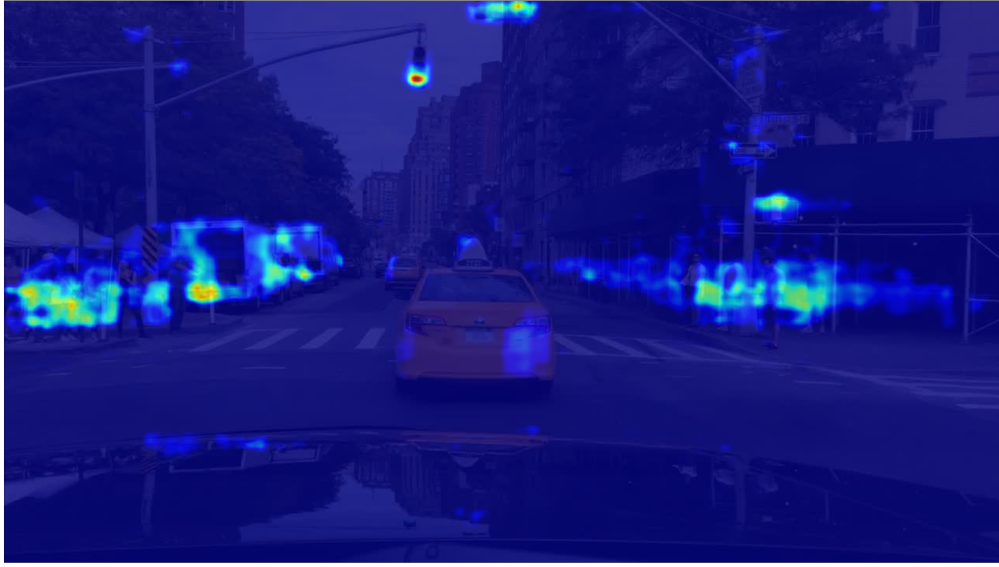}
\includegraphics[width=0.45\linewidth]{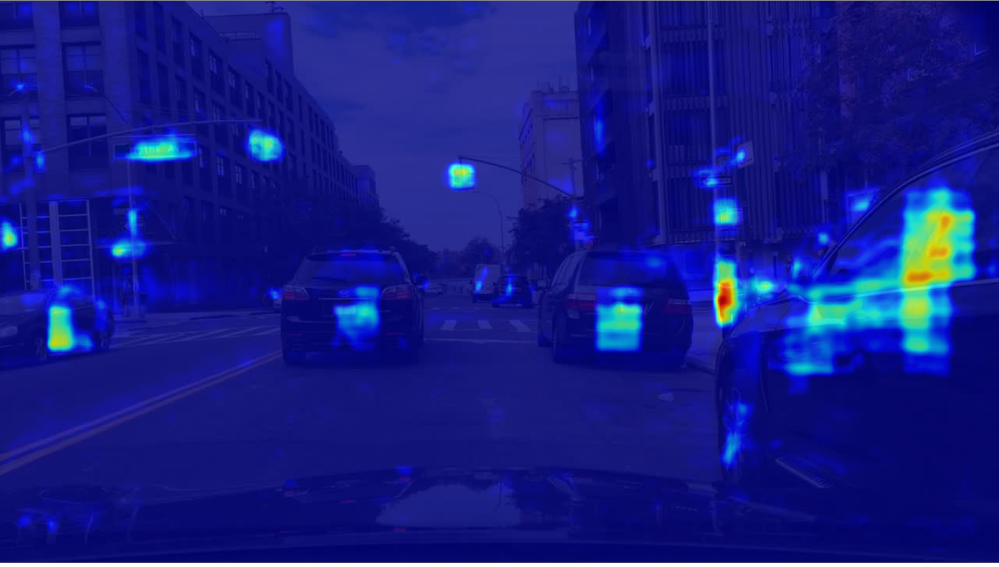}\\
(E)
\end{center}
\end{minipage}
\caption{Saliency maps by grad-CAM: (A) are input images, (B) and (C) are the saliency maps for supervised concepts, and (D) and (E) are the unsupervised concept ones. Concepts of (B) and (C) are ``Obstacle: car'', and ``Traffic light is red''. Note that the green signal turns on in the left-hand side image.}
\label{fig:gradcam}
\end{figure}

Table \ref{tbl:BDD} shows the experimental results of CBM, M-SENN, and CBM-AUC. As shown in this table, the performances of CBM-AUC are better than CBM and M-SENN. Figs.~\ref{fig:CBM-AUC_succeed} and \ref{fig:gradcam} show an example of predictions and saliency maps of some units in the concept layer. As shown in Fig.~\ref{fig:CBM-AUC_succeed}, our model can predict actions and concepts correctly. In addition, from Fig.~\ref{fig:gradcam}, our model can pay attention to areas corresponding to each supervised concept. Furthermore, the unsupervised concepts provide valuable concepts for the action classification. These results indicate the effectiveness of our combination of supervised and unsupervised concepts, as in Sec.~\ref{subsub:cub}.

\subsubsection{Performance for the Different Intermediate Network}

\begin{table}[t]
\centering
{\begin{tabular}{ccccc}
\hline
$\vect{h}$ & $mF1$ & $F1_{all}$ & $mF1_{cpt}$ & $F1_{cpt,all}$ \\
\hline \hline
Inception-v3 & 0.485 & 0.592 &  0.16 & 0.31 \\
Faster RCNN & 0.658 & 0.704 & 0.342 & 0.522 \\
\hline
\end{tabular}}\\
\caption{Performance comparison with respect to changing backbone network.}
\label{tbl:bdd-oia-inc}
\end{table}

\begin{figure}[t]
\centering
\begin{minipage}{1.0\hsize}
\begin{center}
\includegraphics[width=0.45\linewidth]{images/gradCAM_cpt2.png}
\includegraphics[width=0.45\linewidth]{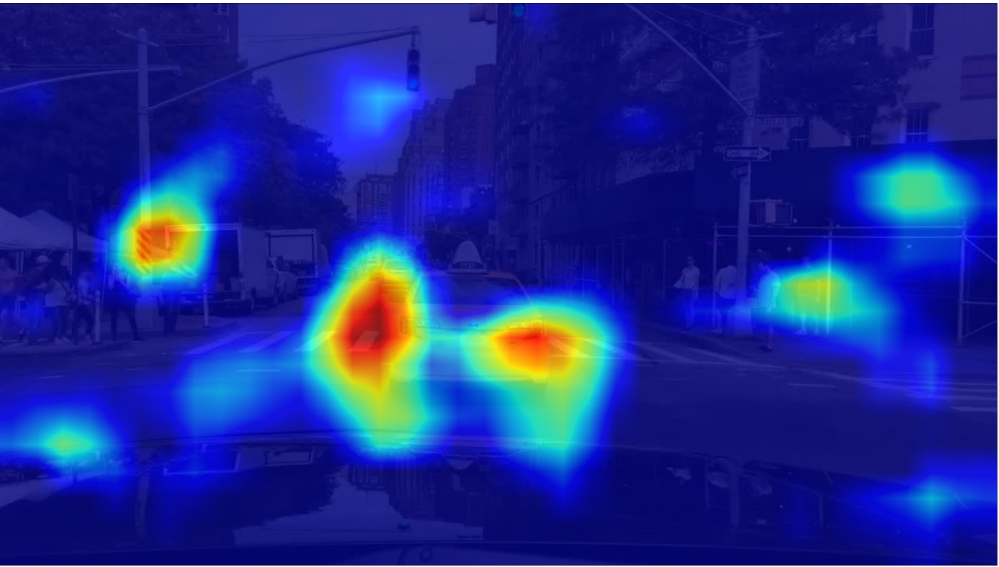}\\
(A)
\end{center}
\end{minipage}
\begin{minipage}{1.0\hsize}
\begin{center}
\includegraphics[width=0.45\linewidth]{images/gradCAM_cpt3.png}
\includegraphics[width=0.45\linewidth]{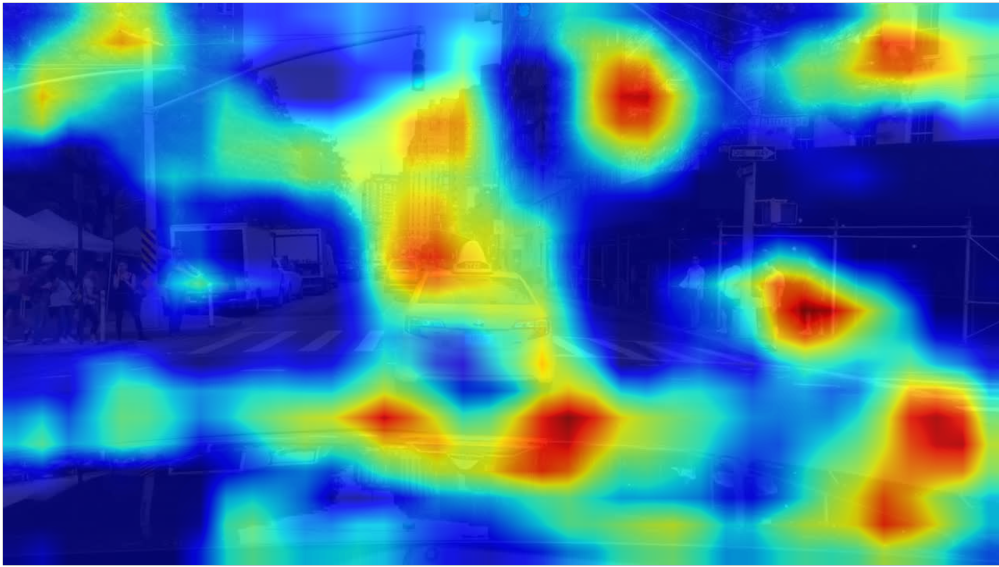}\\
(B)
\end{center}
\end{minipage}
\begin{minipage}{1.0\hsize}
\begin{center}
\includegraphics[width=0.45\linewidth]{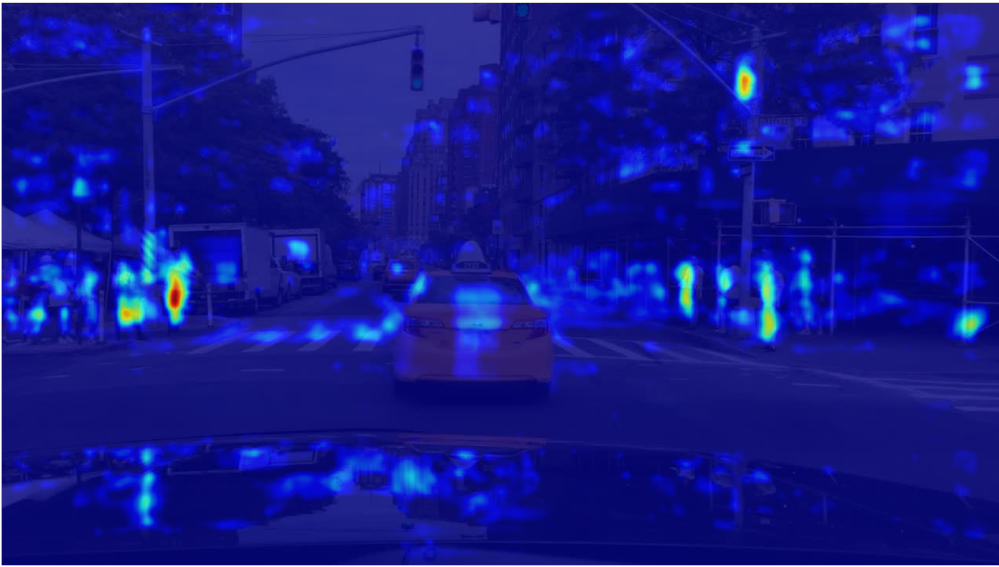}
\includegraphics[width=0.45\linewidth]{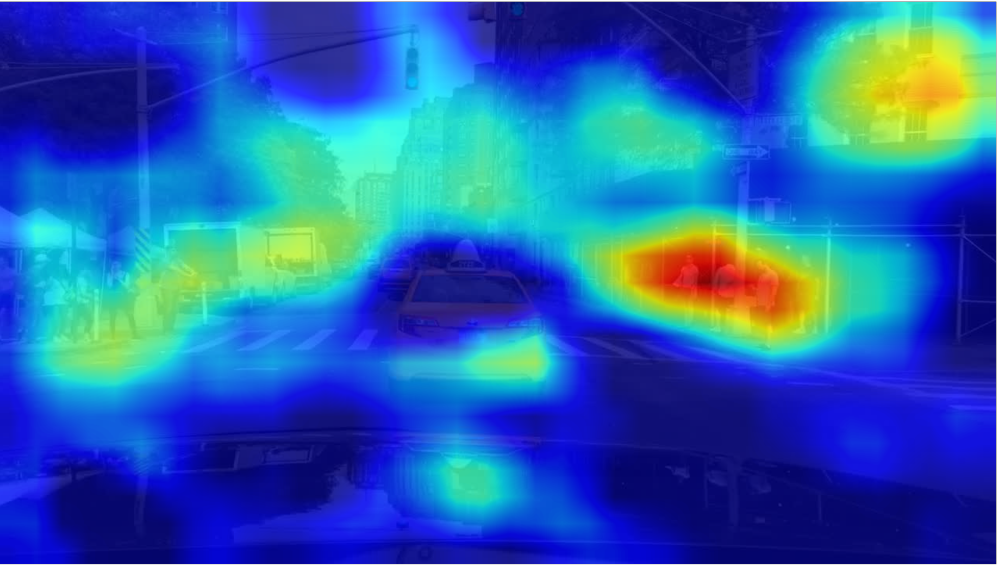}\\
(C)
\end{center}
\end{minipage}
\caption{Comparison of saliency maps when changing $\vect{h}(.)$. Left: Faster RCNN, right: Inception-v3. Each ground truth concept is (A): ``Traffic light is green'', (B): ``Obstacle: car'' and (C): ``Obstacle: person". The same image as the left-hand side of Fig.~\ref{fig:gradcam} (A) is used.}
\label{fig:gradcam_inc}
\end{figure}

To investigate the influence of $\vect{h}(.)$, we replaced $\vect{h}(.)$ from Faster RCNN to Inception-v3. Table~\ref{tbl:bdd-oia-inc} shows the performance comparison. Note that we present results for only four important metrics (i.e.,  $mF1$, $F1_{all}$, $mF1_{cpt}$, and $F1_{cpt,all}$). As shown in this table, the performance when using Inception-v3 was worse than when using the Faster RCNN. Figure \ref{fig:gradcam_inc} shows the saliency maps for some supervised concepts. As shown in this figure, saliency maps using Inception-v3 were not consistent with the semantic meanings. From these results, we consider that the Inception-based model is not suitable for BDD-OIA.

\subsubsection{Performance Comparison with Multi-Task Models}

\begin{table}[t]
\centering
{\begin{tabular}{ccccc}
\hline
Model & $mF1$ & $F1_{all}$ & $mF1_{cpt}$ & $F1_{cpt,all}$ \\
\hline \hline
Global~\cite{xu2020explainable} & 0.392 & 0.601 & 0.180 & 0.331  \\
Local~\cite{wang2019deep} & 0.699 & 0.711 & 0.196 & 0.406  \\
Global+Local~\cite{xu2020explainable} &0.718 & 0.734 & 0.208 & 0.422  \\
CBM-AUC & 0.658 & 0.704 & 0.342 & 0.522 \\
\hline
\end{tabular}}\\
\caption{Performance comparison for different models.}
\label{tbl:BDD_compare}
\end{table}

We also compared the different model architecture. Table~\ref{tbl:BDD_compare} shows the comparison with the multi-task models performed in \cite{xu2020explainable} and \cite{wang2019deep}. ``Global'' is the accuracy when using the simplest global module we also use, ``Local'' uses the local features produced by the RPN and ROI head layers of Fater RCNN, and ``Global+Local'' combines global and local features. Although we cannot compare these results directly since the probability is different\footnote{The models in \cite{wang2019deep,xu2020explainable} compute $\vect{x}\rightarrow(\vect{c},\vect{y})$ while our model computes $\vect{x} \rightarrow \vect{c} \rightarrow y$, where $\vect{x},\vect{c}$ and $y$ correspond to input, concepts and task, respectively.}, concept accuracy of our model was greater than the multi-task models. However, the multi-task models, especially using local features, exhibited better action accuracies than our model. There is a potential to improve the task accuracy of our model by modifying $\vect{h}(.)$, such as utilizing local information; however, this topic remains as our future works.

\section{Discussion}

\begin{figure}[t]
\centering
\begin{minipage}{0.45\hsize}
\begin{center}
\includegraphics[width=1.0\linewidth]{images/result_plot_97-min.jpg}\\
(A)
\end{center}
\end{minipage}
\begin{minipage}{0.45\hsize}
\begin{center}
\includegraphics[width=1.0\linewidth]{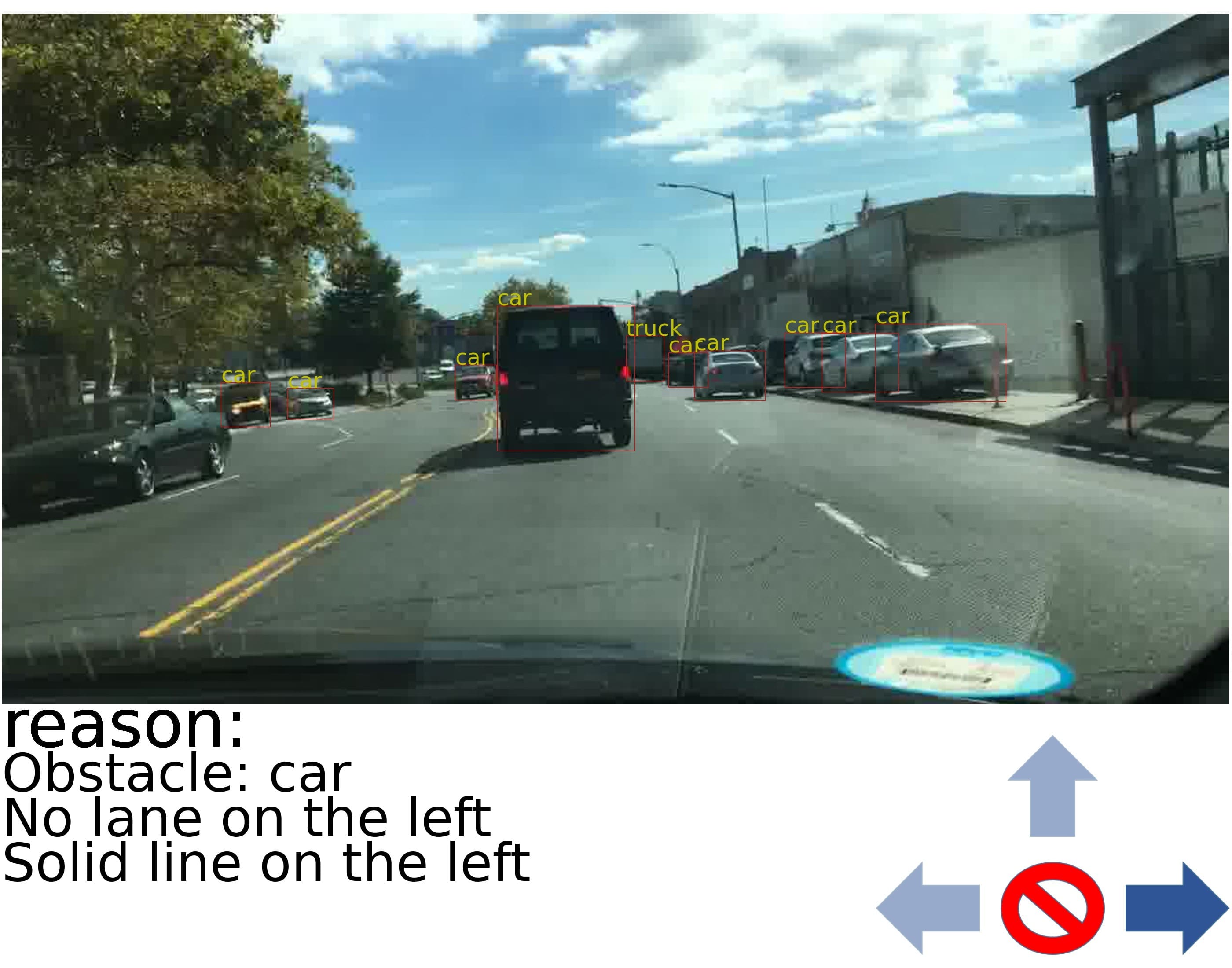}\\
(B)
\end{center}
\end{minipage}
\caption{Examples of ambiguous labels. (A): Ground truth concept is ``Obstacle:car'', and the ground truth actions are ``Stop'', ``Left'', and ``Right''. (B): Ground truth concepts are ``Obstacle:car'' and ``Solid line on the left'', and the ground truth actions are ``Stop'' and ``Right''.}
\label{fig:CBM-AUC_miss}
\end{figure}

It is essential to ``visualize'' where the model is looking at for a reliable prediction. A recent paper~\cite{2021Ashman} showed that the CBM does not look at the region where the concept indicates. However, we found that our model for BDD-OIA seems to look at reasonable areas, as shown in Fig.~\ref{fig:gradcam}. We also suggested that these different results come from the intermediate network $\vect{h}(.)$ shown in Fig.~\ref{fig:gradcam_inc}. 

It is also interesting to observe the saliency maps for unsupervised concepts. As shown in Figs.~\ref{fig:gradcam} (D) and (E), the unsupervised concept is an entangled set of multiple concepts that humans can understand. For example, the right-hand side of Fig.~\ref{fig:gradcam} (D) pays attention to the traffic sign, signals, and front cars. Ideally, these concepts should be able to be extracted separately. How to disentangle human-understandable concepts through unsupervised learning is one of the open problems.

Addressing ambiguous labels is also an important future step. Figure~\ref{fig:CBM-AUC_miss} shows an example. Figures~\ref{fig:CBM-AUC_miss} (A) and \ref{fig:CBM-AUC_succeed} (B) have the same ground truth concept ``Obstacle:car'', whereas, the ground truth action is ``Stop'' for Fig.~\ref{fig:CBM-AUC_succeed} (B) and are ``Stop'', ``Left'', and ``Right'' for Fig.~\ref{fig:CBM-AUC_miss} (B). In both images, ``Stop'' seems reasonable. However, ``Left'' and ``Right'' are difficult to determine from the image in Fig.~\ref{fig:CBM-AUC_miss} (A) alone. In fact, our model outputs ``Stop'' and ``Obstacle:car''. In contrast, the ground truth concepts and actions of Fig.~\ref{fig:CBM-AUC_miss} (B) are ``Obstacle:car'' and ``Solid line on the left'', and ``Stop'' and ``Right''. Our model can output these actions and concepts correctly. However, it may also output a concept that does not seem to be wrong but does not contain the ground truth concepts (``No lane on the left''). Using video as an input may solve this problem. However, many challenges still remain to output appropriate concepts in time series (e.g., the drastic increase in annotation cost).

\section{Conclusion}
In this study, we have proposed Concept Bottleneck Model with Additional Unsupervised Concepts motivated by actual human concepts. To realize this, we have combined CBM and modified SENN to represent explicit and implicit knowledge. We have examined the effectiveness of our model with two datasets, CUB-200-2011 and BDD-OIA, and we found that our model outperformed CBM and modified SENN for all datasets. We also showed that the saliency maps of each concept were consistent with the semantic meaning for BDD-OIA. 

In the future, we plan to adapt the proposed model to video input, improve our model by changing the intermediate layer, and output human-understandable concepts by unsupervised learning. Although opportunities for further research remain, we believe that this approach paves the way from System1 to System 2~\cite{kahneman2011thinking}.

\small{

}
\end{document}